\documentclass{article}
\usepackage{spconf,amsmath,graphicx}
\usepackage{times}
\usepackage{epsfig}
\usepackage{graphicx}
\usepackage{amsmath}
\usepackage{amssymb}

\usepackage{makecell}
\usepackage[table ]{ xcolor}
\definecolor{mygray}{gray}{.9}

\usepackage{subfigure}
\usepackage{algorithm}  
\usepackage[noend]{algpseudocode}

\title{SM+: Refined Scale Match for Tiny Person Detection}
\name{Nan Jiang, Xuehui Yu, Xiaoke Peng, Yuqi Gong and Zhenjun Han\thanks{The corresponding author is Zhenjun Han.}}
\address{University of Chinese Academy of Sciences, Beijing, China}

\begin{document}
%
\maketitle
\begin{abstract}
Detecting tiny objects (\emph{e.g.}, less than $20\times20$ pixels) in large-scale images is an important yet open problem. Modern CNN-based detectors are challenged by the scale mismatch between the dataset for network pre-training and the target dataset for detector training. In this paper, we investigate the scale alignment between pre-training and target datasets, and propose a new refined Scale Match method (termed SM+) for tiny person detection. SM+ improves the scale match from image level to instance level, and effectively promotes the similarity between pre-training and target dataset. Moreover, considering SM+ possibly destroys the image structure, a new probabilistic structure inpainting (PSI) method is proposed for the background processing. Experiments conducted across various detectors show that SM+ noticeably improves the performance on TinyPerson, and outperforms the state-of-the-art detectors with a significant margin.
\end{abstract}
\section{Introduction}
\label{sec:intro}
\vspace{-2mm}
Person detection is an important topic in the computer vision area. It has wide applications including surveillance \cite{Collins2000ASF}\cite{Haritaoglu2000W4RS}, driving assistance \cite{Dollr2009PedestrianDA} and maritime quick rescue \cite{Yu2020ScaleMF}, \emph{etc.} The research of detectors \cite{lin2017feature}\cite{Cai2018CascadeRD}\cite{Pang2019LibraRT}\cite{lin2017focal}\cite{Tian2019FCOSFC} has achieved significant progress with the rapid development of data-driven deep convolutional neural networks (CNNs). 
However, the detectors perform poorly when detecting tiny objects with few pixels (\emph{e.g.}, less than $20\times20$ pixels), such as traffic signs \cite{Lu2018TrafficSD}, persons in aerial images \cite{Yu2020ScaleMF}, \emph{etc.}

To better exploit the CNN-based detectors, a large number of person datasets \cite{Everingham2009ThePV}\cite{lin2014microsoft}\cite{Kuznetsova_2020} for detection with human manual annotations have been proposed and made publicly available. However, datasets for specific object detection, such as tiny person detection \cite{Yu2020ScaleMF}, are not as large as other counterparts \cite{deng2009imagenet}\cite{lin2014microsoft}, due to the cost of collecting and annotating the data. With the insufficient data for a specific application, an alternative way is to pre-train a model on the extra-large datasets (\emph{e.g.}, ImageNet \cite{deng2009imagenet}, COCO \cite{lin2014microsoft}), and then fine-tune the model on a task-specific dataset.

However, a new question arises: Could we take better advantage of existing large datasets for a task-specific application, particularly when object sizes significantly differ between the datasets? SM algorithm\cite{Yu2020ScaleMF}, Random Scale Match (RSM) and Monotone Scale Match (MSM), gave simple yet effective ways. With a sampled scale factor, the SM algorithm directly resizes the images and aligns the scale distribution of pre-training dataset to that of the target dataset.
The SM algorithm, with image-level scaling, is merely a simple approximation for scale match by simply regarding the average size of all objects in an image as the size of the image, where there may be many labeled objects with multi-scales.

\begin{figure}[tb]
\begin{center}
   \includegraphics[width=1\linewidth]{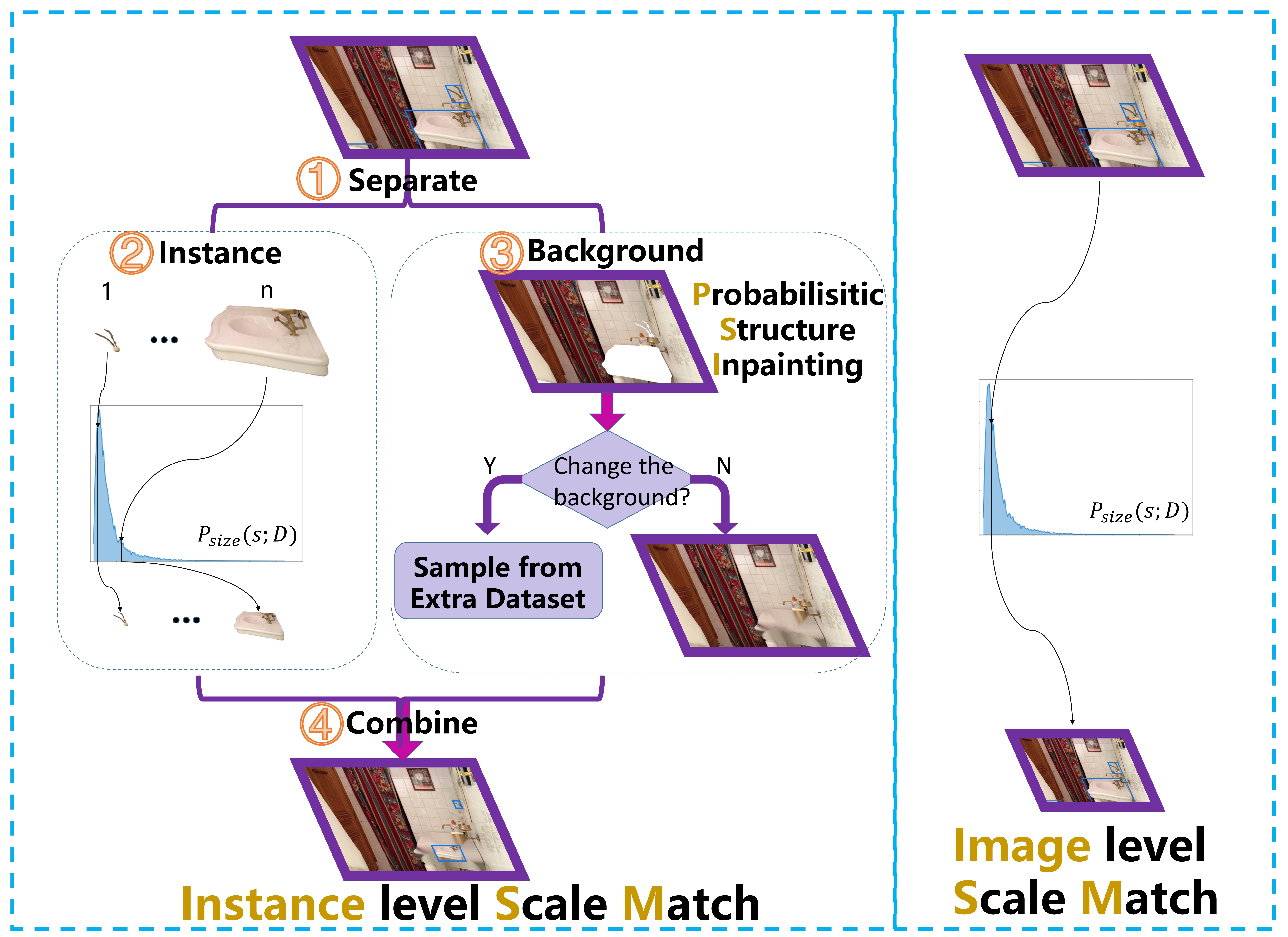}
\end{center}
\vspace{-6mm}
   \caption{The illustration of the difference between Image-level SM and Instance-level SM. While SM only considers the whole image, SM+ focuses on every instance. The instance-level approach achieves scale match in a \emph{finer} level,
   which mainly consists of four steps: (1) Separation, (2) Instance processing, (3) Background processing, and (4) Combination.}
\label{fig:different}
\end{figure}

In this paper, we propose a newly refined SM method (SM+), in which we transform the scale distribution of pre-training dataset by instance-level scaling instead of resizing the whole image. Intuitively, compared with the vanilla SM algorithm, SM+ is a finer-scale scaling and alleviates the uncertainty and inaccuracy caused by the approximation introduced in the SM algorithm. The differences between SM and SM+ are illustrated in Fig. \ref{fig:different}. SM+ algorithm separates the image into two parts: the annotated instances and their background. The instances are utilized for the instance-level scale match, while the images are destroyed by some instance-shaped holes (background). However, the traditional method \cite{Bertalmo2001NavierstokesFD} blurs the images by direct inpainting the holes and generates some unreal images, leading to the performance drop of the pre-trained model. To solve this problem, a probabilistic structure inpainting (PSI) method is further proposed to dynamically inpaint the images by suppressing the image blur and preserving context consistency around the holes. Compared with state-of-the-art detectors on TinyPerson, SM+ algorithm leads to significant performance improvement in AP. The main contributions of our work include:

1. We comprehensively analyze the scale information of TinyPerson, and propose a new refined scale match method, dubbed as SM+, which achieves better scale distribution alignment by finer-scale scaling.

2. We propose the probabilistic structure inpainting for the SM+ algorithm. PSI can effectively inpaint the images.

3. The proposed SM+ algorithm improves the detection performance over the state-of-the-art detectors with a large margin. Codes will be available upon acceptance.

\section{Methodology}
\setlength{\parskip}{1pt}
\vspace{-2mm}
\subsection{Scale Match}
\vspace{-2mm}
We define the object size as the square root of its area: $s(G_{ij})=\sqrt{w_{ij}h_{ij}}$ where $G_{ij}$ denotes the $j$-th bounding box of $i$-th image $I_i$, and $w_{ij}, h_{ij}$ are the width and height of the bounding box, respectively. 

Given an extra dataset $E$ where the probability density function of object size $s$ is $P_{size}(s;E)$ and a target dataset $D$ where the probability density function is $P_{size}(s;D)$, our goal is to apply a scale transformation $T$ on $E$, such that their probability distributions of object size can be well matched. This is corresponding to,

\vspace{-1mm}
\begin{small}\begin{equation}\label{distribution}
P_{size}(s;T(E)) 
\approx P_{size}(s;D).
\end{equation}\end{small}\vspace{-1mm}Image-level method leaves lots of room for improvement. To this end, we propose the refined scale match (SM+), which focuses on instance-level scale match and achieves more desirable results than image-level match \cite{Yu2020ScaleMF}. 

\vspace{-4mm}
\subsection{SM+: Refined Scale Match}
\vspace{-2mm}

The whole procedure is shown in Fig. \ref{fig:different}. In the following, we present the details of each part.

\noindent\textbf{Part I. Extraction and Separation:} Pre-training dataset requires ground-truth annotations for instance segmentation. According to the mask annotation, each picture participating in the training is separated into the background and foreground. In order to get the finer foreground, we adopt the matting method \cite{He2011AGS} to make the outline of instances smoother. Because the stored form of mask annotation is boundary points and edges, using such annotations directly makes the outline of the foreground jagged. After separation, we get a proper instance mask and an incomplete background. Then the two parts are processed separately.

\noindent\textbf{Part II. Instance Scale Histogram Match:} On the ground of target dataset annotation, a discrete scale histogram $H$ is established to approximate the scale probability density function of target dataset $P_{size}(s; D_{train})$, which is rectified to pay less attention to the long-tail part of scale distribution. 
In $H$, $K$ represents the number of bins in scale histogram, $R[k]^-$ and $R[k]^+$ are size boundaries of $k$-th bin. 
For every separated instance, we use the size of the corresponding bounding box as its scale representation $s$. First, we sample an index of bin with respect to the probability of $H$. Then we sample a target scale $\hat{s}$ based on a uniform probability distribution, whose min and max size equal to $R[k]^-$ and $R[k]^+$, respectively. Finally, we transform the instance according to the ratio of $\hat{s}$ to $s$. It can be defined by the affine transformation matrix,
\begin{small}\begin{equation}     
A
=
\left[                
  \begin{array}{ccc}   
    r & 0 & t_x\\  
    0 & r & t_y\\  
    0 & 0 &1\\
  \end{array}
\right],                
\end{equation}\end{small}where $r$ denotes the scale variance, $t_{x}$ and $t_{y}$ denote the coordinate shift in $x$-axis and $y$-axis, respectively. 

\begin{figure}[tb]
\begin{center}
  \includegraphics[width=1\linewidth]{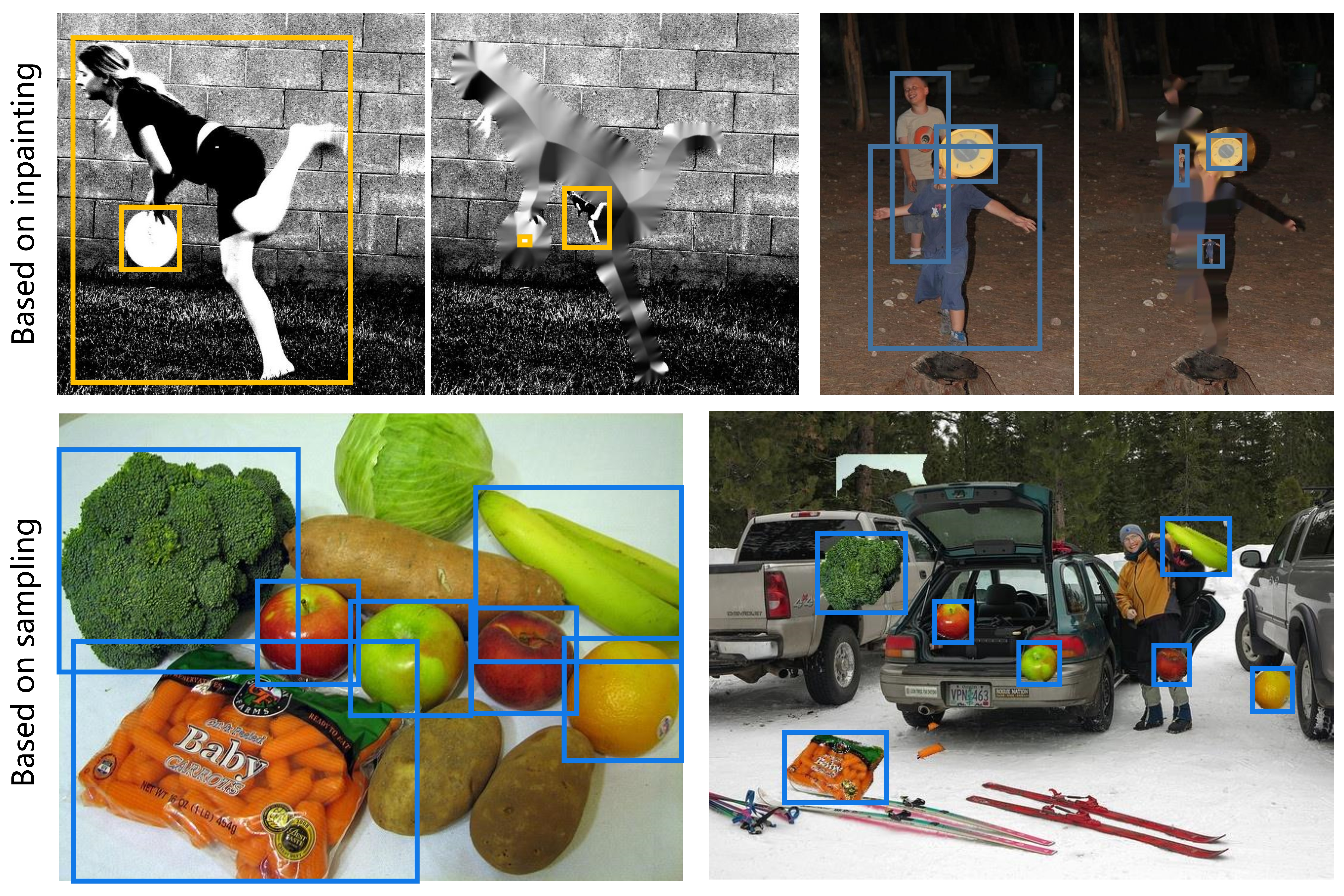}
\end{center}
\vspace{-6mm}
  \caption{Background based on inpainting (\textbf{top}) vs. Background based on new sampling (\textbf{bottom}). The inpainting method might not repair some artifacts
  , but changing the background does not cause this problem. 
  (Best viewed in color.)}
\label{fig:example}
\end{figure}

\noindent\textbf{Part III. Probabilistic Structure Inpainting:} For such a background with an instance-shaped hole on it, we first adopt the inpainting method \cite{Bertalmo2001NavierstokesFD} to fill in the blank area of the background inspired by InstaBoost \cite{Fang2019InstaBoostBI}. In practice, however, the effect of the traditional inpainting method can be very poor because the object is reduced to a very small size. 
In order to alleviate the image structure loss caused by instance-level scale match, we introduce extra background to make up for the distortion of the image. This also raises the question that the context information of the object will be completely different from before. To some extent, it will confuse network learning. Therefore, a hyper-parameter $p$ is predefined to determine whether a new background is needed. We use it to find a trade-off between the two kinds of background. If the random number is greater than $p$, we will sample a new image from the pre-training dataset as background. On the contrary, we still use the inpainting background. It should be noted that the label of the new image will not participate in training. 

\noindent\textbf{Part IV. Combination:} After getting the final background and reasonable instance, we paste the transformed instances on the corresponding position in background according to annotation. 
Adjusted images can be visualized in Fig. \ref{fig:example}. 

\vspace{-3mm}
\subsection{Discussion}

\vspace{-2mm}
In order to prove the effectiveness of SM+, we use the Jensen-Shannon divergence \cite{Lin1991DivergenceMB} to quantitatively measure the similarity between distributions.
Here, $p(x)$ and $q(x)$ denote probability distribution of a discrete random variable $x$. Both $p(x)$ and $q(x)$ sum up to 1, and $p(x) > 0$ and $q(x) > 0$ for any $x$ in $X$. 
Kullback-Leibler divergence\cite{Kullback1951ONIA} \begin{small}
$D_{KL}(p(x),q(x))$\end{small} is defined in Eq. (\ref{KL})
\begin{small}\begin{equation}\label{KL}
D_{KL}(p(x)||q(x))=\sum_{x\in X}p(x)ln\frac{p(x)}{q(x)}.\end{equation}\end{small}Therefore, we can get the formulation of Jensen-Shannon divergence \begin{small}
$D_{JS}(p(x),q(x))$
\end{small} from Eq. (\ref{JS})  

\begin{small}\begin{equation}\begin{aligned}\label{JS}
D_{JS}(p(x)||q(x))=\left.\sum_{x\in X}[\frac{1}{2}D_{KL}(p(x)||\frac{p(x)+q(x)}{2})
\right.
\\
\phantom{=\;\;}
\left.+\frac{1}{2}D_{KL}(q(x)||\frac{p(x)+q(x)}{2})].\right.\end{aligned}\end{equation}\end{small}

\setlength{\tabcolsep}{7mm}
\begin{table}[tb]
\begin{center}
\begin{tabular}{l|c}
\Xhline{1.2pt}
 $T$ & \begin{small}
 $D_{JS}(P_{size}(s;T(E))||P_{size}(s;D))$
 \end{small} \\
\Xhline{1.2pt}
\rowcolor{mygray}RSM & 0.0091\\
RSM+ &0.0020\\
\rowcolor{mygray}MSM & 0.0133\\
MSM+ &0.0013\\

\Xhline{1.2pt}
\end{tabular}
\end{center}
\vspace{-6mm}
\caption{The similarity between the scale distributions aligned by different methods. A smaller similarity score denotes the two distributions are closer. 
$D$ represents TinyPerson, $E$ represents COCO, and $T$ denotes the transformation conducted by the scale match method.}
\label{tab:JS}
\end{table}

According to Tab. \ref{tab:JS}, 
the JS divergence between scale distribution transformed by SM+ algorithm is less than that transformed by SM algorithm. 
The proposed SM+ more effectively bridges the gap between the scale distribution of pre-training dataset and target dataset.

\section{Experiment}
\vspace{-2mm}
\subsection{Dataset}\label{subsection1}
\vspace{-2mm}
\indent The experiments are conducted in two datasets: COCO and TinyPerson. \textbf{COCO} involves 80 categories of objects. 
\textbf{TinyPerson} is a tiny object detection dataset collected from high-quality videos and web pictures. TinyPerson contains 72,651 annotated human objects with a low resolution on visual effect in total 1, 610 images. 
The size of most annotated objects in TinyPerson is less than $20\times20$ pixels.

\vspace{-4mm}
\subsection{Comparison with the state-of-the-art methods}\label{subsection3}

\setlength{\tabcolsep}{9.5pt}
\begin{table*}[t]

    \begin{center}
    \small
    \begin{tabular}{l|c|c|c|c|c||c|c}
        \Xhline{1.2pt}
        Method & $AP^{tiny1}_{50}$ & $AP^{tiny2}_{50}$ & $AP^{tiny3}_{50}$ & $AP^{tiny}_{50}$ & $AP^{small}_{50}$& $AP^{tiny}_{25}$& $AP^{tiny}_{75}$\\
        \Xhline{1.2pt}
        \rowcolor{mygray}FCOS \cite{Tian2019FCOSFC}                        & 0.99 & 2.82& 6.20& 3.26  & 20.19 & 13.28 & 0.14\\
        Adaptive RetinaNet\cite{lin2017focal}                              & 27.08& 52.63& 57.88& 46.56 & 59.97 & 69.60 & 4.49 \\
        \rowcolor{mygray}Faster RCNN-FPN \cite{lin2017feature}           & 30.25& 51.58& 58.95& 47.35 & 63.18 & 68.43 & 5.83\\
        \Xhline{1.2pt}
        Faster RCNN-FPN-RSM \cite{Yu2020ScaleMF}	        &33.91	&55.16	&62.58	&51.33	&66.96	&71.55	&6.46 \\
        \rowcolor{mygray}Faster RCNN-FPN-RSM+ (ours)  &33.74 &55.32 &62.95 &51.46 &66.68 &72.38 &6.62\\
        Faster RCNN-FPN-MSM	\cite{Yu2020ScaleMF}        &33.79	&55.55	&61.29	&50.89	&65.76	&71.28	&6.66 \\
        \rowcolor{mygray}Faster RCNN-FPN-MSM+ (ours)  &\textbf{34.20} &\textbf{57.60} &\textbf{63.61} &\textbf{52.61} &\textbf{67.37} &\textbf{72.54} &\textbf{6.72}\\
        \Xhline{1.2pt}
    \end{tabular}
     \end{center}
    \vspace{-6mm}
    \caption{Comparisons in terms of $AP$s (\%). Larger $AP$ means better performance.  $AP_{50}^{tiny}$, $AP_{50}^{tiny1}$,  $AP_{50}^{tiny2}$, $AP_{50}^{tiny3}$,  $AP_{50}^{small}$ reflect the performance of object size in range [2, 20], [2, 8], [8, 12], [12, 20], [20, 32], respectively. The \textbf{Bold} indicates the best performance.}
    \vspace{-2mm}
\label{tab:baseline ap}
\end{table*}

\vspace{-2mm}
In Tab. \ref{tab:baseline ap}, Faster RCNN-FPN-MSM+, Faster RCNN-FPN pre-trained with our proposed MSM+, produces state-of-the-art results in all AP evaluations. The comparison well demonstrates that our method is effective for tiny object detection.

\setlength{\tabcolsep}{11mm}
\begin{table}[t]
\small
\begin{center}
\begin{tabular}{l|c}
\Xhline{1.2pt}
 Pre-training Dataset & $AP^{tiny}_{50}$ ($\uparrow$) \\
\Xhline{1.2pt}
\rowcolor{mygray}ImageNet & 47.35\\
COCO800& 49.76\\
\rowcolor{mygray}RSM (COCO) &51.33\\
RSM+ (COCO) &51.46\\
\rowcolor{mygray}MSM (COCO) &50.89\\
MSM+ (COCO) &\textbf{52.61}\\

\Xhline{1.2pt}
\end{tabular}
\end{center}
\vspace{-6mm}
\caption{Comparisons of $AP^{tiny}_{50}$ with \textbf{Faster RCNN-FPN}. Compared with SM algorithm, SM+ algorithm shows how to perform a better pre-training at a deeper level.}
\label{table3}
\end{table}

\setlength{\tabcolsep}{11mm}
\begin{table}[t]
\small
\begin{center}
\begin{tabular}{l|c}
\Xhline{1.2pt}
 Pre-training Dataset  & $AP^{tiny}_{50}$ ($\uparrow$) \\
\Xhline{1.2pt}
\rowcolor{mygray}ImageNet & 46.56\\
COCO800& 45.03\\
\rowcolor{mygray}RSM (COCO) &48.48\\
RSM+ (COCO) &50.59\\
\rowcolor{mygray}MSM (COCO) &49.59\\
MSM+ (COCO) &\textbf{51.25}\\
\Xhline{1.2pt}
\end{tabular}
\end{center}
\vspace{-6mm}
\caption{Comparisons of $AP^{tiny}_{50}$ on \textbf{Adaptive RetinaNet}. SM+ algorithm achieves consistent performance improvement with the one-stage detector.}
\label{table4}
\end{table}

\vspace{-4mm}
\subsection{Analysis}\label{subsection4}
\vspace{-2mm}
\setlength{\parskip}{1pt}
\noindent\textbf{Pre-training Strategy:} As shown in Tab. \ref{table3}, we compare SM+ with various pre-training strategies including ImageNet, COCO800, RSM \cite{Yu2020ScaleMF} and MSM \cite{Yu2020ScaleMF}. The COCO800 means that we control the size of images in (800, 1333) as input and use different anchor settings for each of the two training stages. For COCO we use the original as input. Scale match based methods are applied to COCO dataset. Faster RCNN-FPN is used as the detector. 
First, compared with ImageNet, using COCO800 for pre-training can improve performance with a proper anchor setting since the TinyPerson contains much smaller objects than COCO. 

Considering the person scale distribution of TinyPerson, RSM and MSM can achieve higher accuracy. Furthermore, SM+ can effectively match the scale of COCO to that of TinyPerson and improve detection accuracy. For example, RSM+ outperforms 0.13 point over RSM in $AP^{tiny}_{50}$. Moreover, using the monotone function proposed in \cite{Yu2020ScaleMF}, we get MSM+, which gains an improvement of about 1.72\% over MSM. 

\noindent\textbf{Detector-agnostic:} In order to further validate the efficiency of the proposed approach, one-stage detector Adaptive RetinaNet is also chosen as baseline. In Tab. \ref{table4}, the improvement in one-stage detector is more than that in two-stage detector. RSM+ improves $AP^{tiny}_{50}$ by 2.11 points. MSM+ also improves $AP^{tiny}_{50}$ by 1.66 points, and $MR^{tiny}_{50}$ by 1.30 points. 

The performance improvement of one-stage detector is significantly greater than that of two-stage detector.
In Tab. \ref{table3} and Tab. \ref{table4}, the consistent improvement on both kinds of detector demonstrates that the proposed refined scale match (SM+) is detector-agnostic, which can be effectively used for different kinds of detectors.

\noindent\textbf{Probabilistic Structure Inpainting (PSI):} 
We note that simply aligning scale distributions of pre-training dataset and target dataset at instance level does not improve performance since the image structure is destroyed. 
SM+ involves significantly zooming out objects, where the inpainting method might not be effective in repairing image. This will cause some artifacts, and damage the image structure as shown at the top of Fig. \ref{fig:example}. In contrast, PSI allows instances to be pasted on another background image. In this case, the resulting image will not have artifacts as shown in the bottom of Fig. \ref{fig:example}. 
To validate the effect of PSI, we include a baseline without the background change (w/o PSI) in Tab. \ref{psi}. We show that this baseline drops performance dramatically.
We believe the unrealistic image structure and artifact pattern make network over-fitting, leading to undesirable results.

Moreover, replacing the background in PSI might be regarded as data augmentation. Thus, we further conduct experiments to validate whether the performance gain is from data augmentation. To study this, we include an experiment: directly copy and paste objects on a new background image without scaling its size. We introduce two baselines, CP and CP+.
CP means we crop all the instances and paste them on a new image background, but the original annotations of the new image will not be used during pre-training.
CP+ means both newly pasted objects and original annotated objects are used for training.
In Tab. \ref{table7}, the two baselines achieve similar results and slightly surpass COCO. However, they are lower than MSM+ (COCO). This indicates that replacing the background can only bring limited improvement but it is not the mechanism of our SM+. The effectiveness of SM+ comes from achieving a finer distribution alignment at instance level and better preserving the structure of the images. 

In addition, we also validate the effect of $p$ in PSI and show the results. We observe that a moderate probability ($p$=0.4) can achieve a trade-off between image structure loss and semantic loss.

\setlength{\tabcolsep}{11mm}
\begin{table}
\small
\begin{center}
\begin{tabular}{l|c}
\Xhline{1.2pt}
 Pre-training Strategy & $AP^{tiny}_{50}$ ($\uparrow$) \\
\Xhline{1.2pt}
\rowcolor{mygray}RSM+ (w/o PSI)&50.12\\
RSM+ &51.46\\
\rowcolor{mygray}MSM+ (w/o PSI) &50.69\\
MSM+ &52.61\\

\Xhline{1.2pt}
\end{tabular}
\vspace{-6mm}
\end{center}
\caption{Ablation study on PSI. It is not enough to align the distribution at instance level without considering the background. SM+ can achieve the desired effect with PSI.}
\label{psi}
\end{table}

\setlength{\tabcolsep}{11mm}
\begin{table}
\small
\begin{center}
\begin{tabular}{l|c}
\Xhline{1.2pt}
 Pre-training Dataset & $AP^{tiny}_{50}$ ($\uparrow$) \\
\Xhline{1.2pt}
\rowcolor{mygray}COCO & 49.96\\
CP (COCO) &50.66\\
\rowcolor{mygray}CP+ (COCO) &50.46\\
MSM (COCO) &50.89\\
\rowcolor{mygray}MSM+ (COCO) &52.61\\

\Xhline{1.2pt}
\end{tabular}
\end{center}
\vspace{-6mm}
\caption{Effect of different methods. CP (COCO) and MSM (COCO) both achieve limited performance improvements.}
\label{table7}
\end{table}

\vspace{-2mm}
\section{Conclusion}
\vspace{-2mm}
The scale information for better pre-training is further investigated in this paper. Scale Match only focuses on the image-level match and thus limits the feature representation learning for detectors. In this paper, we propose a novel method named Refined Scale Match (SM+). SM+, a much finer scale match strategy, aligns scale distributions of pre-training dataset and target dataset at instance level, yielding a more effective and suitable matched dataset. Moreover, in order to alleviate the loss caused by aligning distribution at instance level, an effective method, referred to as probabilistic structure inpainting (PSI), is further proposed. PSI effectively balances the information loss between image structure and semantics. Thorough experimental results verified the superiority of the proposed method over other state-of-the-art methods. 

The relative size between two datasets is also very important for tiny object detection, which will be further investigated in the future.

\bibliographystyle{IEEEbib}
\bibliography{strings,refs}

\end{document}